\documentclass[journal]{IEEEtran}
\usepackage{graphicx} 
\usepackage{mathrsfs}
\usepackage{amsthm,amsmath,amssymb,lipsum}
\usepackage{multirow}
\usepackage{booktabs}
\usepackage{times}
\usepackage{epsfig}
\usepackage{graphicx}
\usepackage{amsmath}
\usepackage{amssymb}
\usepackage{color}
\usepackage{subfig}
\usepackage{url}
\usepackage[numbers]{natbib}
\usepackage{setspace}
\usepackage[colorlinks,
            linkcolor=blue,
            anchorcolor=blue,
            citecolor=blue]{hyperref}
\usepackage{xspace}
\newcommand{\eg}{\textit{e.g.}\xspace}
\newcommand{\ie}{\textit{i.e.}}
\usepackage{siunitx}
\usepackage{bbding}

\begin{document}
\title{Near-infrared Image Deblurring and Event Denoising with Synergistic Neuromorphic Imaging}
\author{Chao~Qu,
        Shuo~Zhu,
        Yuhang~Wang,
        Zongze~Wu,
        Xiaoyu~Chen,
        Edmund~Y.~Lam,
        and~Jing~Han
\thanks{Chao~Qu, Yuhang~Wang, Zongze~Wu, Xiaoyu~Chen, and Jing~Han are with Jiangsu Key Laboratory of Spectral Imaging and Intelligent Sense, Nanjing University of Science and Technology, Nanjing 210094, China (e-mail: chao.qu@njust.edu.cn; yhgwang@njust.edu.cn; zongze@njust.edu.cn; 115104000466@njust.edu.cn; eohj@njust.edu.cn).}
\thanks{Shuo~Zhu and Edmund~Y.~Lam are with the Department of Electrical and Electronic Engineering, The University of Hong Kong, Pokfulam, Hong Kong SAR, China (e-mail: zhushuo@hku.hk; elam@eee.hku.hk).}
\thanks{Chao~Qu and Shuo~Zhu contributed equally to this work.}
\thanks{Jing~Han and Edmund~Y.~Lam are the corresponding authors.}
}

\maketitle
\begin{abstract}
The fields of imaging in the nighttime dynamic and other extremely dark conditions have seen impressive and transformative advancements in recent years, partly driven by the rise of novel sensing approaches, \eg, near-infrared (NIR) cameras with high sensitivity and event cameras with minimal blur. However, inappropriate exposure ratios of near-infrared cameras make them susceptible to distortion and blur. Event cameras are also highly sensitive to weak signals at night yet prone to interference, often generating substantial noise and significantly degrading observations and analysis. Herein, we develop a new framework for low-light imaging combined with NIR imaging and event-based techniques, named synergistic neuromorphic imaging, which can jointly achieve NIR image deblurring and event denoising. Harnessing cross-modal features of NIR images and visible events via spectral consistency and higher-order interaction, the NIR images and events are simultaneously fused, enhanced, and bootstrapped. Experiments on real and realistically simulated sequences demonstrate the effectiveness of our method and indicate better accuracy and robustness than other methods in practical scenarios. This study gives impetus to enhance both NIR images and events, which paves the way for high-fidelity low-light imaging and neuromorphic reasoning.

\end{abstract}


\IEEEpeerreviewmaketitle
\section{Introduction}
\IEEEPARstart{I}{maging} in low-light conditions has diverse applications, from micro biomedical analysis to macro astronomy observation, and is especially crucial for ensuring the safety of emerging autonomous driving at night~\cite{li2021low,liu2021benchmarking}. Under low-light environments, due to limited sensor sensitivity, we have to trade off signal-to-noise ratio for motion blur compensation or vice versa, posing a significant problem for these practical applications. Despite impressive advancements with new hardware and innovative algorithms, enhancing low-light imaging remains a formidable challenge.
\begin{figure}[t]
    \centering
    \includegraphics[width=0.98\linewidth]{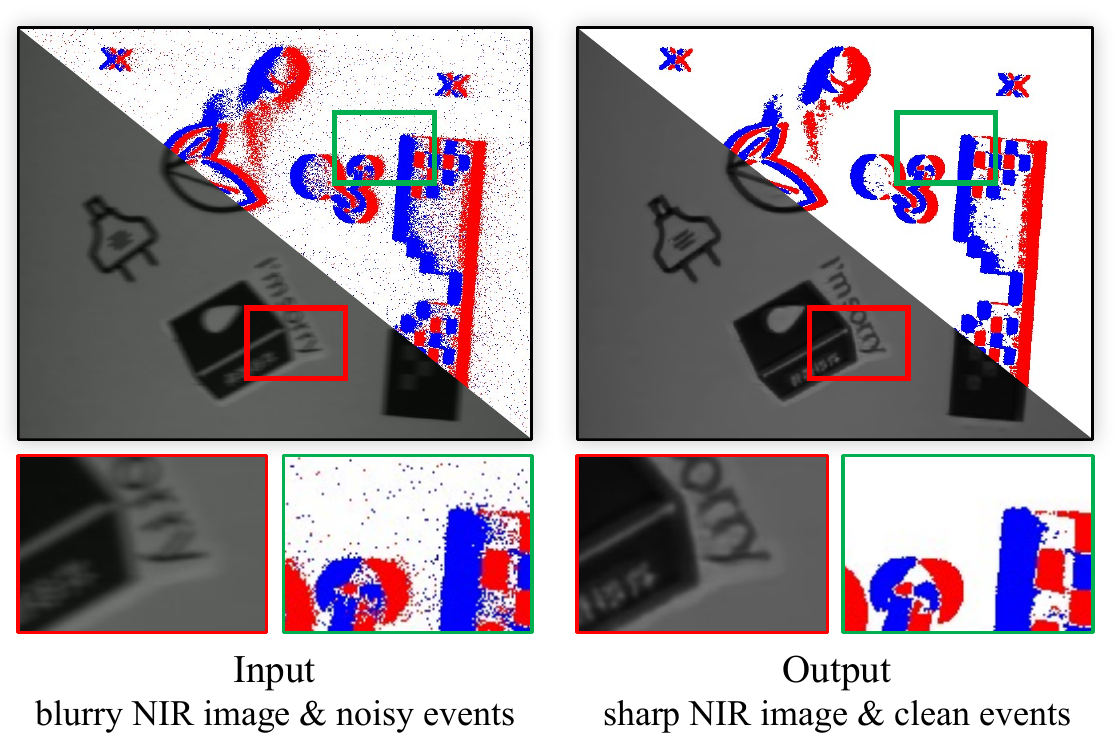}
    \vspace{-0.5ex}
    \caption{Comparison examples of shared view with NIR image and events. A single NIR and event camera is often accompanied by motion blur and noise in low-light environments. Our synergistic scheme can reconstruct the sharp NIR image and clean events.}
    \label{fig1_teaser}
\end{figure}
Near-infrared (NIR) imaging has a unique sensitivity that projects distant objects with clarity, making it particularly useful for night vision applications~\cite{niu2023nir}. NIR imaging has garnered significant attention due to its biomedical imaging, environmental monitoring, and surveillance applications. However, traditional NIR imaging systems often struggle with improper settings, leading to degraded image quality and hindering subsequent analysis with motion blur and artifacts~\cite{geng2024event}. Additionally, the NIR band exists beyond the human visible spectrum, enabling the use of NIR light that is invisible to humans~\cite {li2020spectrum}. Therefore, the NIR images may present structures inconsistent with visible light images, primarily due to the inherent differences in optical properties of the target material across the two spectral regions~\cite{xu2024nir}. Event cameras, capturing visual information as asynchronous events rather than traditional frames, are central to this evolution. This approach mimics the human visual system, enabling superior temporal resolution and improved performance in dynamic scenes. Computational neuromorphic imaging (CNI) with event cameras offers advantages like minimal motion blur and enhanced dynamic range compared to conventional methods. However, the sparse and noisy nature of low-light events leads to a distribution that differs significantly from that in normal light. Events in low-light scenes have distinct distributions prone to various types of interference and increased noise~\cite{gallego2020event,zhu2024computational}. Although event cameras excel in capturing high-speed motion and low-light conditions, they also present unique challenges, particularly in noise reduction. As shown in Fig.~\ref{fig1_teaser}, the NIR image suffers from blur, and events are polluted with significant noise in low-light environments. A sharp NIR image and clean events are essential for downstream imaging tasks.

NIR imaging and the CNI paradigm are well characterized, creating a new demand to synergize them for advanced imaging performance. We can surpass these limitations by harnessing each attractive property of NIR and event cameras with computational imaging techniques to enhance clarity, detail, and fidelity in NIR images and events. In this work, we focus on fusion consistency and leveraging modal inconsistency, aiming to circumvent its adverse effects on combinations of NIR images and visible events in a new and efficient paradigm. To address the vulnerabilities and limitations of NIR and CNI applications, this paper advocates a novel approach to NIR image deblurring and event denoising by integrating two divergent sensors synergistically. 

To address these challenges, the paper contributes by demonstrating:

\begin{itemize}
    \item We design a synergistic neuromorphic imaging approach for NIR image deblurring and event denoising, leveraging NIR imaging and visible events to enhance performance through complementary strengths.
    \item We developed a new framework to effectively integrate cross-modal features of NIR images and visible events via spectral consistency and higher-order interaction.
    \item We build a new cross-spectral deblurring and denoising (CSDD) dataset containing paired NIR images and event streams. Extensive qualitative and quantitative comparisons demonstrate the effectiveness and robustness of the proposed method.
\end{itemize}

\section{Related Work}
\label{sec:related}
\subsection{The NIR imaging and Image Deblurring}
NIR imaging can capture infrared light at wavelengths between $700$ and $1200$ nanometers, which is invisible to the human eye. These longer wavelengths exhibit unique scattering and absorption characteristics compared to visible light, allowing NIR imaging to achieve higher signal-to-noise ratios and capture more information, especially in low-light or other challenging environments \cite{jin2022darkvisionnet,cheng2023mutually}. As a result, NIR imaging is increasingly applied in computational imaging and computer vision, showing great potential in applications such as image denoising \cite{yan2013cross,xu2024nir}, image fusion \cite{li2020spectrum,yang2023detail}, material classification \cite{salamati2009material}, and scene category recognition \cite{brown2011multi}. However, NIR cameras often have low frame rates, leading to motion blur. Traditional deblurring methods, which rely on deconvolution algorithms \cite{schmidt2013discriminative,xu2014inverse} or specific priors \cite{chan1998total,dong2017blind}, face limitations in generalization and effectiveness. Recently, deep learning approaches have shown improvements in image reconstruction quality by learning the complex mapping between blurry and sharp images \cite{tao2018scale,zhou2022lednet,ren2023multiscale,mao2023intriguing,gao2024efficient,liu2024motion}. Additionally, some works have applied the high temporal resolution of event cameras to image deblurring, achieving better results than traditional frame-based methods \cite{pan2019bringing,xu2021motion,sun2022event,zhou2023deblurring,nakabayashi2023event,yang2023event,zhang2024crosszoom,sun2025motion}. Unlike these approaches, this paper focuses on the collaborative enhancement of NIR and event cameras, aiming to achieve NIR images deblurring through cross-spectral information fusion.

\subsection{The CNI paradigm and Event Denoising}
Unlike conventional cameras that capture full images at a fixed rate determined by an external clock, event cameras are retinomorphic devices that trigger an event once the observed logarithmic intensity variation exceeds a preset threshold and record the fast dynamics more effectively~\cite{gallego2020event}. This makes them ideally suited for advanced sensing and offers significant potential for computer vision and computational imaging in challenging scenarios~\cite{chakravarthi2024recent,zhu2024computational}. The high sensitivity with logarithmic response raises exciting possibilities for low-light imaging, offering superior performance to scientific detectors and enabling high-resolution imaging ability and weak signal detection~\cite {cabriel2023event}. However, event cameras with logarithmic response are particularly susceptible to different types of interference, leading to increased noise levels. Therefore, denoising of events becomes a crucial preprocessing step for downstream tasks~\cite{delbruck2008frame,9720086}. Research on event denoising is primarily categorized into filtering-based methods and learning methods. Filtering-based methods rely on manual priors to design discriminative models for noise removal, employing techniques like density distinction with motion continuity~\cite{feng2020event,zhang2023neuromorphic}, and motion compensation within the spatiotemporal domain~\cite{8244294}. However, the effectiveness of these priors can vary with different signal/noise event distributions, which may limit their denoising accuracy. Learning-based methods for event denoising have been studied and used recently, such as EDnCNN \cite{baldwin2020event}, EventZoom \cite{duan2021eventzoom}, AEDNet \cite{fang2022aednet}, LED \cite{duan2024led}, etc. Given the growing need for sensitive and dynamic processing capabilities, CNI-informed techniques that provide high-quality events with reduced noise are essential for driving progress in complex applications.

\begin{figure*}[t]
    \setlength{\abovecaptionskip}{6pt}
    \setlength{\belowcaptionskip}{-0.2cm}
    \centering
    \includegraphics[width=0.98\linewidth]{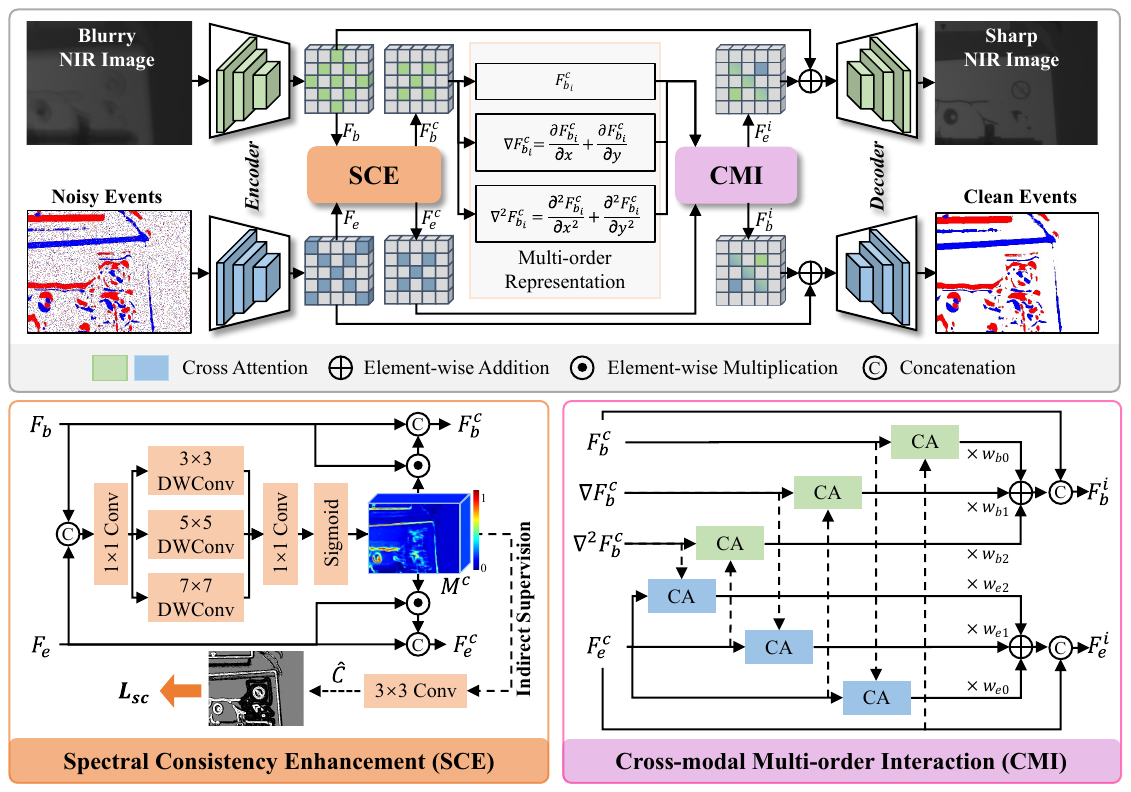}
    \vspace{-0.5ex}
    \caption{Architecture of the proposed MDEDNet, which consists of a dual-branch subnetwork, a spectral consistency enhancement (SCE) module, and a cross-modal multi-order interaction (CMI) module. MDEDNet takes a blurry NIR image and noisy events as input, and through collaborative enhancement, simultaneously outputs a sharp image and clean events.}
    \label{fig:overview}
\end{figure*}

\section{Methodology}
\label{sec:method}

\subsection{Problem Formulation}
We first revisit the fundamental models for motion deblurring and event denoising. Next, we establish a unified task framework that is based on NIR and neuromorphic imaging, to integrate the complementary advantages of synergistic imaging and simultaneously achieve NIR motion deblurring and event denoising. For this unified task, we propose a dual-modal synergistic scheme in the following section.

\noindent\textbf{Motion Deblurring.} The blurry NIR image \(B\), can be formulated as the average of the latent intensity images \( I(t) \) over the exposure interval \( T \) 
\begin{align}\label{eq:image_blur}
B = \frac{1}{T} \int_{0}^{T} I(t) \, dt,
\end{align}
where \( T \) denotes the exposure time. The motion deblurring aims at recovering a sharp NIR image \( S \) from the blurred image \( B \)
\begin{align}\label{eq:motion_deblurring}
S = \mathcal{B}^{-1} (B, E),
\end{align}
where \( \mathcal{B}^{-1} \) represents the NIR motion deblurring operation assisted by the event streams \(E=\{ (x_i, y_i, t_i, p_i) \}_{i=0}^T \). Here, \( x_i \) and \( y_i \) indicate the pixel positions, \( t_i \) represents the time of the \( i \)-th event, and \( p_i \in \{-1, +1\} \) denotes polarity, representing brightness change directions.

\noindent\textbf{Event Denoising.}
Due to hardware limitations, events are inevitably contaminated by noise \( N \), especially in low-light conditions
\begin{align}\label{eq:event_noise}
E_n = E_c + N,
\end{align}
where \( E_c \) represents the noise-free events, and \( E_n \) is the noisy events. The goal of event denoising is to recover the noise-free event \( E_c \) from the noisy event \( E_n \)
\begin{align}\label{eq:event_denoising}
E_c = \mathcal{E}^{-1}(E_n, I),
\end{align}
where \( \mathcal{E}^{-1} \) denotes the event denoising operation assisted by NIR images \(I\).

\noindent\textbf{Unified Task.}
The unified task is simultaneous NIR image deblurring and event denoising, which can be expressed as
\begin{align}\label{eq:unified_task}
S, E_c = \mathcal{BE}^{-1}(B, E_n),
\end{align}
where \(\mathcal{BE}^{-1}\) represents a unified NIR motion deblurring and event denoising operation based on synergistic neuromorphic imaging.

\subsection{Synergistic Scheme}
We propose a novel NIR and event cooperative enhancement network, MDEDNet, which simultaneously achieves image motion deblurring and event denoising by exploring the mutual compensation between degraded dual-modal (blurred NIR images and noisy events). Fig. \ref{fig:overview} illustrates the architecture of the proposed MDEDNet, which consists of a dual-branch subnetwork, a spectral consistency enhancement (SCE) module, and a cross-modal multi-order interaction (CMI) module. Each branch of the subnetwork adopts a multi-scale encoder-decoder structure, with encoding and decoding layers at each scale consisting of two Res Blocks.

\subsubsection{Spectral Consistency Enhancement}
Since NIR imaging and event imaging capture information in different spectral bands (\ie, NIR and visible wavelength), structural discrepancies may occur between the events and NIR images in certain scenes. Ignoring this structural inconsistency during the fusion of these heterogeneous data may introduce artifacts, thereby reducing the accuracy of dual-modal cooperative enhancement. To address this issue, we propose a Spectral Consistency Enhancement (\textbf{SCE}) module, which assigns pixel-level weights by predicting a spectral consistency map \( M^c \) to suppress inconsistent structures between NIR and event features.

As illustrated in Fig. \ref{fig:overview}, the NIR feature \( F_{b} \in \mathbb{R}^{C \times H \times W} \) and event feature \( F_{e} \in \mathbb{R}^{C \times H \times W} \) are used as inputs to estimate the spectral consistency map \( M^c \in \mathbb{R}^{C \times H \times W} \). Then, \( M^c  \) is multiplied with \( F_{b} \) and \( F_{e} \) respectively to enhance common features. This process is described as follows
\begin{align}\label{eq:sce2}
M^c &= \text{SCE}(F_{b}, F_{e}), \nonumber\\
F^c_{b} &= Concat\big([F_{b}, F_{b} \odot M^c ]\big), \nonumber\\
F^c_{e} &= Concat\big([F_{e}, F_{e} \odot M^c ]\big),
\end{align}
where \(\odot\) denotes the element-wise multiplication operation. \(F^c_{b}\) represent the NIR consistency enhancement features. \(F^c_{e}\) represent the event consistency enhancement features. To improve the estimation accuracy of the spectral consistency map \( M^c \), we use large-kernel depth-wise convolutional layers \cite{howard2017mobilenets} to capture richer structural information. Specifically, we integrate depth-wise convolutional layers with various kernel sizes (\ie, \(3 \times 3\), \(5 \times 5\), \(7 \times 7\)) in the SCE and then use a \(1\times1\) convolution to aggregate features under different receptive fields. Finally, the aggregated features are nonlinearly mapped to the spectral consistency map \(M^c\) through the Sigmoid function according to the relative importance of each pixel location.

Considering SCE struggles to naturally capture structural differences in the feature dimensions between NIR images and events, we introduce explicit supervision to the SCE for generating a high-accuracy \( M^c \). Inspired by \cite{jin2022darkvisionnet}, we calculate the structure consistency \( C \in \mathbb{R}^{1 \times H \times W} \) between the visible image corresponding to the events and the NIR image. The definition for \( C \) is as follows
\begin{align}\label{eq:sce3}
C = \frac{1}{2}(1 - {\rm S}^{v})(1 - {\rm S}^{n})+ {\rm S}^{v} \times {\rm S}^{n},
\end{align}
where \( {\rm S}^{v} \) denotes the edges of the visible image and \( {\rm S}^{n} \) denotes the edges of the NIR image. The pixel value of  \( C \) varies from $0$ to $1$, with $0$ indicating inconsistent structure and $1$ indicating consistent structure, which is suitable for the supervision of SCE. Considering the dimensional difference between \( C \) and \( M^c \), we decode the potential structure consistency \( \widehat{C} \in \mathbb{R}^{1 \times H \times W}\) from the high-dimensional \( M^c \) by a \(3 \times 3\) convolution and constrain it through a spectral consistency loss \(L_{sc}\), which can be expressed as
\begin{align}\label{eq:loss_consistency}
L_{sc} = {|C - \widehat{C}|}_2 = {|C - \text{Conv}(M^c)|}_2.
\end{align}

This approach can indirectly provide effective supervision for the high-dimensional feature space of NIR and events, guiding the SCE module to generate a physically meaningful spectral consistency map \( M^c \), providing reliable and accurate spatiotemporal features for subsequent cross-modal multi-order interaction. 

\subsubsection{Cross-modal Multi-order Interaction}
\begin{figure}[t]
    \centering
    \setlength{\abovecaptionskip}{5pt}
    \setlength{\belowcaptionskip}{-0.2cm}
    \includegraphics[width=0.98\linewidth]{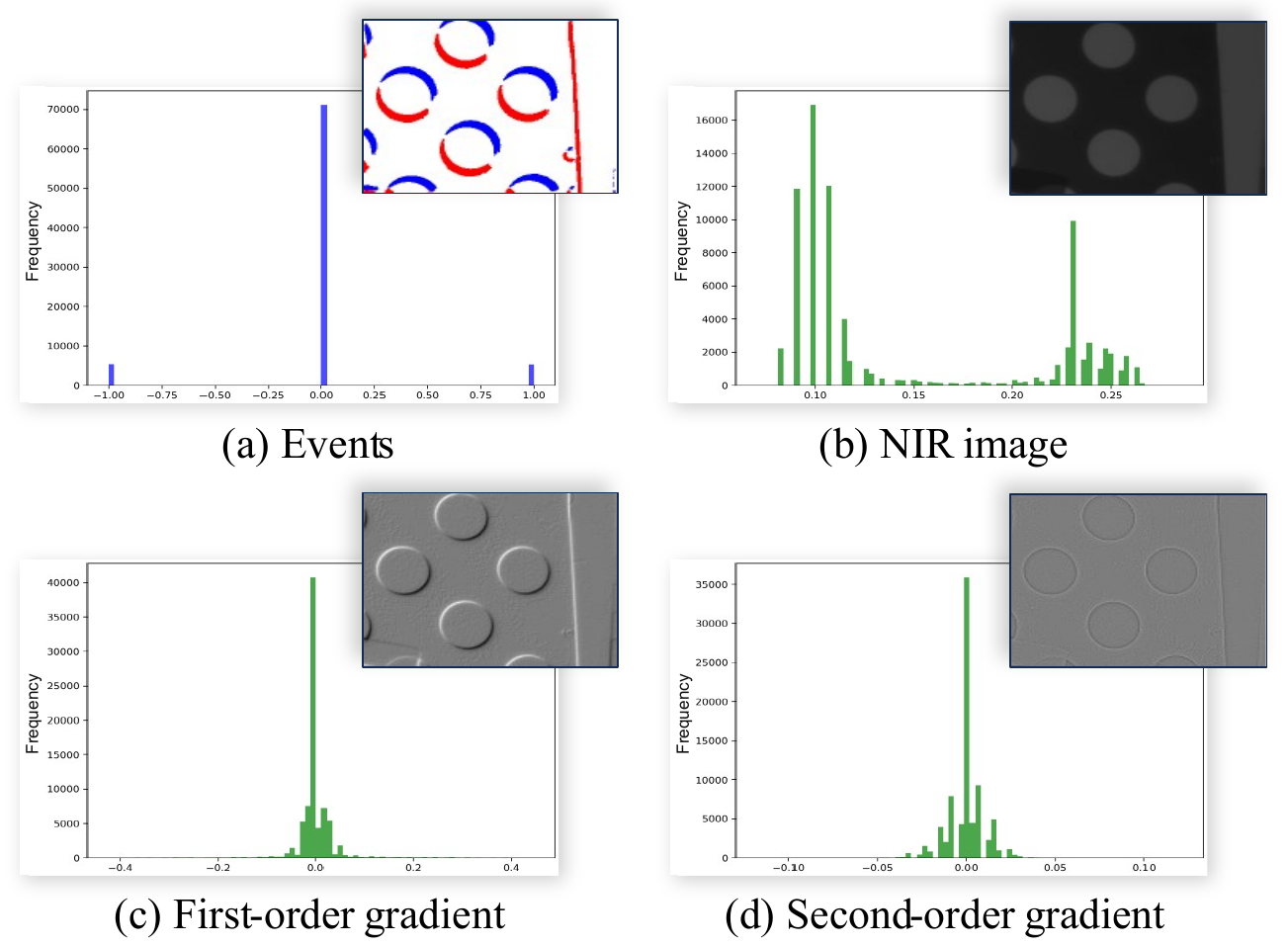}
    \vspace{-0.5ex}
    \caption{(a)-(d) represent events, NIR image, first-order and second-order gradient, respectively, along with their histograms.}
    \label{fig:image_grad_event}
\end{figure}
The significant differences in sparsity and signal-to-noise ratio exist between NIR images and events. Previous studies have not thoroughly investigated the relationships between these two modalities and typically apply concatenates \cite{jiang2020learning} or basic attention mechanisms \cite{sun2022event} to fuse the two modal features. Such interaction methods often struggle to achieve satisfactory results when dealing with complex scenarios. 

To comprehensively analyze the correlation between events and the NIR image, we visualize the events, NIR image, multi-level gradient signals of the image, and their corresponding histograms. First, both the event and image gradient signals reveal the texture features of the scene. As illustrated in Fig. \ref{fig:image_grad_event}, the event camera captures brightness changes in the scene, making edges more likely to trigger events, while flat regions have a lower probability of triggering events. Image gradients can precisely capture the texture details in the image, and these texture features exhibit a spatial distribution similarity with the events. Furthermore, both the event signals and image gradient signals exhibit significant sparsity. The histogram in Fig. \ref{fig:image_grad_event} shows that most pixels in the original image signal have non-zero values, while the most pixels in the image gradients and events are zero. This sparsity indicates that the multi-order gradient signals of sharp images and the event information have substantial collaborative potential in modal interaction.

Based on the above analysis, we propose a cross-modal multi-order interaction (\textbf{CMI}) module that achieves fine-grained fusion between dual modalities through interaction between event features and multi-order image gradients. Specifically, the first-order gradient effectively captures large-scale structural details, helping to identify prominent edges in the scene \cite{zhuang2022underwater}. The second-order gradient, by capturing higher-order edge information at smaller scales, enables more sensitive recognition of subtle texture variations \cite{yang2023detail}. We therefore integrate image features, first-order gradients, and second-order gradients to mitigate the limitations of a single gradient. 

As illustrated in Fig. \ref{fig:overview}, in the feature interaction from events to images, we first compute the first- and second-order gradients of the consistency image features \(F^c_{b}\). Next, we generate the \(Q_{b0}\), \(Q_{b1}\), and \(Q_{b2}\) matrices from the multi-order gradient features, while generating the corresponding \(K_e\) and \(V_e\) matrices from the consistency event features \(F^c_{e}\). We then perform lookups of event features using multi-order gradient features, followed by weighted summation for fusion. Finally, we concatenate the fused features with the event features to obtain the interaction features \(F^i_{e}\). This process can be represented as:
\begin{align}\label{eq:sce1}
F_{e0} &= CA(Q_{b0}, K_e, V_e) \nonumber\\
F_{e1} &= CA(Q_{b1}, K_e, V_e) \nonumber\\
F_{e2} &= CA(Q_{b2}, K_e, V_e) \\
F_e^i = Concat\big([F_e^c, (&w_{e0} \ast F_{e0} + w_{e1} \ast F_{e1} + w_{e2} \ast F_{e2})]\big),\nonumber
\end{align}
where \(CA(\cdot)\) denotes the cross attention operation \cite{vaswani2017attention}, \(F_{ej}\) represents the queried event features corresponding to the \(j\)-th order gradient, and \(w_{ej}\) is the learnable weight for the \(j\)-th order gradient. The process of information interaction from events to images mirrors that from images to events.

By integrating the multi-order features of the image with the spatiotemporal features of the event, we achieve a deep-level interaction between different modalities, effectively reducing image blur and minimizing event noise.

\subsubsection{Loss Function}
The motion deblurring task is supervised by the \(l1\) norm between a predicted sharp image \(\widehat{S}\) and the ground truth sharp image \(S\), as follows
\begin{align}\label{eq:loss_deblur}
L _ {md} = ||\widehat {S} - S||_{1}.
\end{align}

The event denoising task is supervised by the \(l2\) norm  between the predicted event \(\widehat{E}_{c}\) and the ground truth clean event \(E_{c}\), which can be defined as
\begin{align}\label{eq:loss_denoise}
L _ {ed} = | | \widehat{E}_{c} - E_{c}| |_{2}.
\end{align}

Thus, the total loss function for MDEDNet is defined as 
\begin{align}\label{eq:loss_total}
L_{total} = \lambda_1 L _ {md}  + \lambda_2 L_{ed} + \lambda_3 L_{sc},
\end{align}
where \(\lambda_1\), \(\lambda_2\) and \(\lambda_3\) represent the weights of each loss.

\section{CSDD Dataset}
\begin{figure}[t!]
    \centering
    \setlength{\abovecaptionskip}{5pt}
    \setlength{\belowcaptionskip}{-0.3cm}
    \includegraphics[width=0.98\linewidth]{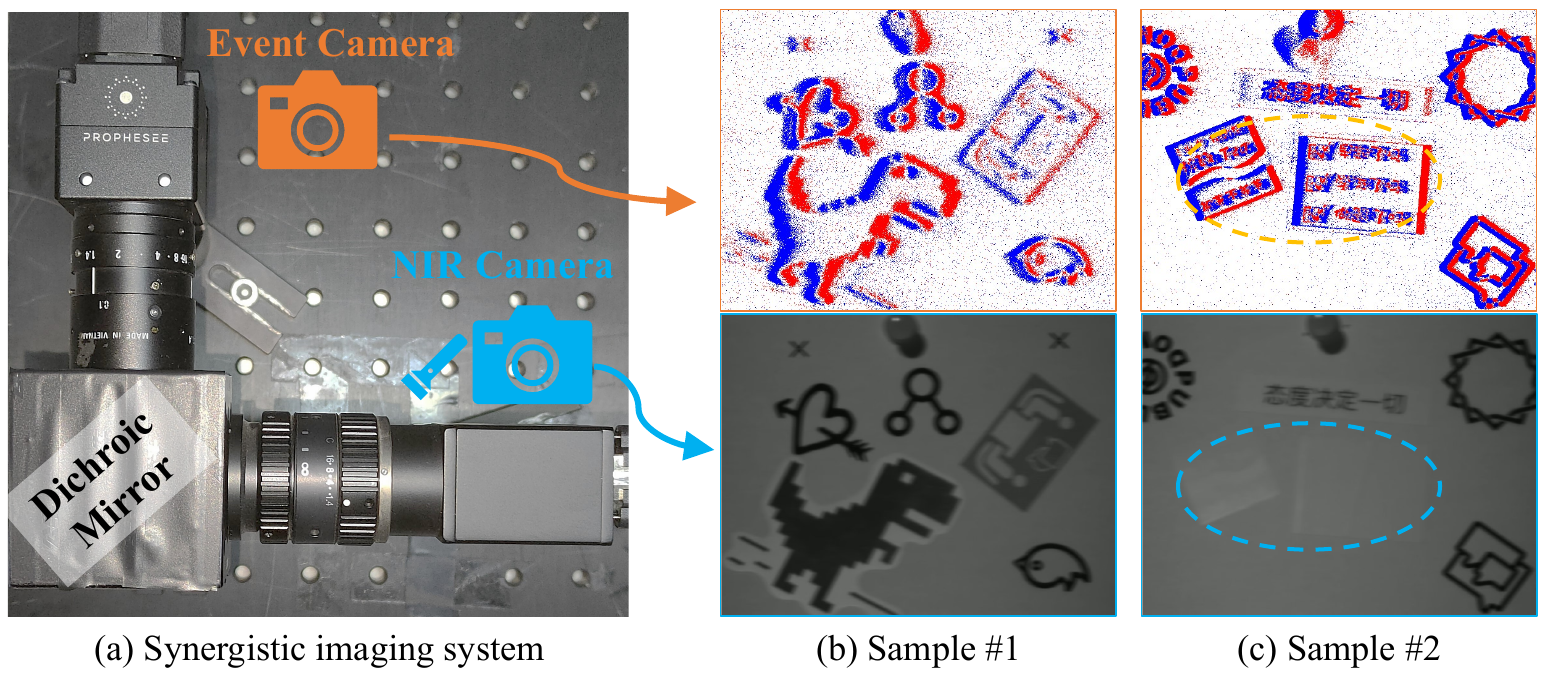}
    \caption{(a) The synergistic imaging system with a NIR camera and an event camera. (b) and (c) Two imaging samples with raw collected NIR images and corresponding noisy events. Notably, the dashed line area in sample \#2 highlights spectral inconsistencies between the NIR and the events.}
    \label{fig:camera_samples}
\end{figure}
\begin{figure*}[t]
    \setlength{\abovecaptionskip}{3pt}
    \setlength{\belowcaptionskip}{-0.0cm}
    \centering
    \includegraphics[width=0.98\linewidth]{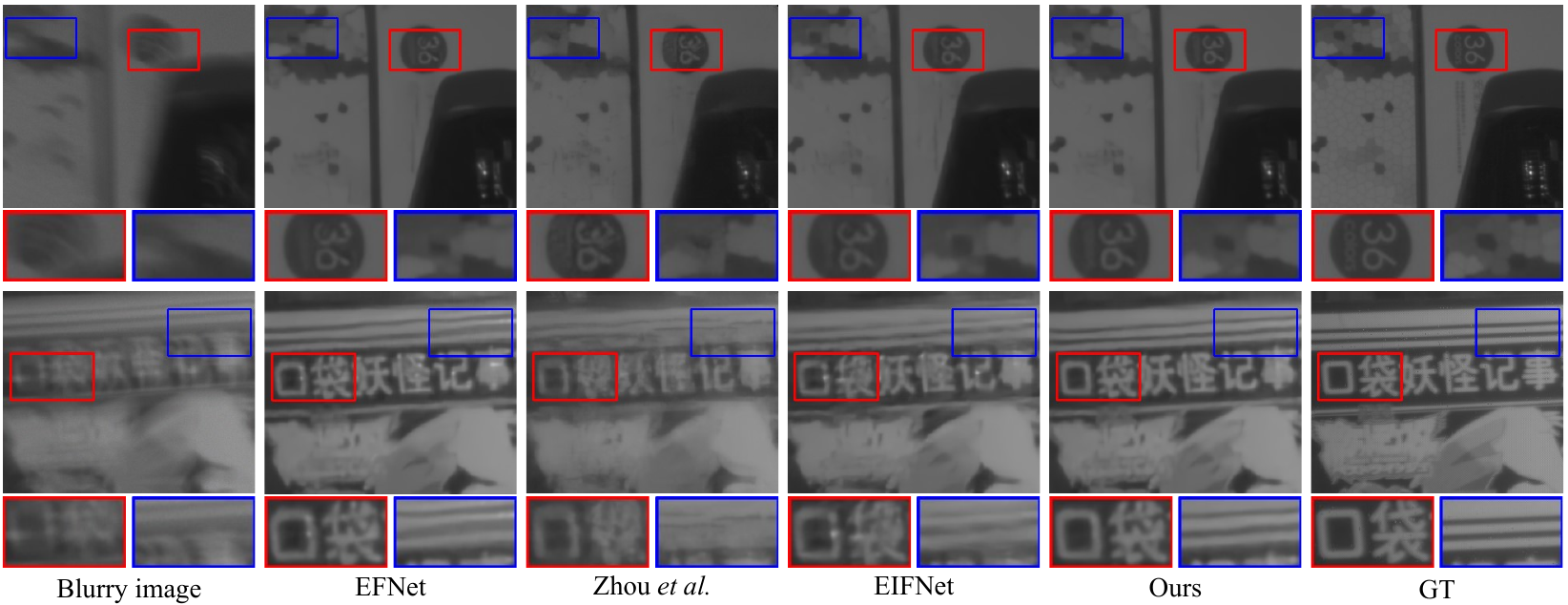}
    \caption{Comparison results of our deblurring method with others on the synthetic dataset. }
    \label{fig:deblur_syn}
\end{figure*}
\begin{figure*}[t]
    \setlength{\abovecaptionskip}{3pt}
    \setlength{\belowcaptionskip}{-0.1cm}
    \centering
    \includegraphics[width=0.98\linewidth]{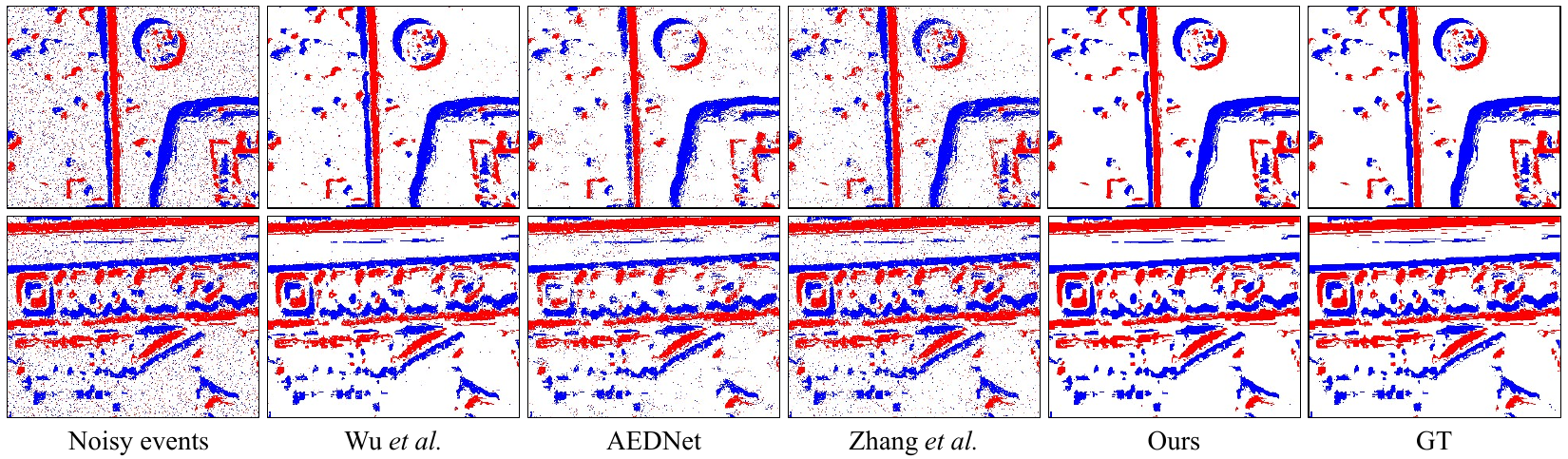}
    \caption{Comparison results of our denoising method with others on the synthetic dataset.}
    \label{fig:denoise_syn}
\end{figure*}
To our knowledge, there is currently no publicly available dataset for NIR and event camera cooperative tasks. This prompted us to establish a new dataset for cross-spectral deblurring and denoising (CSDD), which includes paired blurry-sharp NIR images and corresponding noisy event streams.

To simultaneously capture NIR images and events, we built a synergistic imaging system with an event camera (Prophesee EVK5, \(1280\times720\))  and a NIR camera (HIKROBOT MV-CA016-10GM, \(1440\times1080\)), as shown in Fig. \ref{fig:camera_samples}. To ensure minimum spatial parallax between the NIR camera and the event camera, a dichroic mirror with a cutoff wavelength of \SI{805}{nm} is used to separate the visible light and NIR information reflected from the target. Additionally, an \SI{850}{nm} NIR illuminator is provided for the NIR camera. 

We collect $45$ sequences comprising NIR image sequences and event streams, covering various illumination conditions and spectral response scenarios. During data acquisition, we fix the target object on a motorized rail to achieve precise motion control, enabling the imaging system to capture NIR sequences and event streams simultaneously. To obtain event streams with different noise levels, we conduct imaging under both normal and low-light conditions. Furthermore, we take the impact of spectral inconsistency and select target scenes with different visible light and NIR spectral responses. Since the two cameras provide different visual modalities, calibration in both spatial and temporal domains is crucial to ensure alignment between the collected NIR images and event data. Similar to previous work \cite{zhou2023deblurring}, we adopt an approximate temporal synchronization strategy, writing a script to simultaneously trigger the capturing programs of the two cameras. We then estimate the homography matrix between the two cameras by manually selecting feature points, achieving geometric alignment of the NIR image sequences with the event stream. In our experiments, we crop the central region of both NIR images and event streams to \(640\times512\) pixels.
\section{Experiments}
\label{sec:results}
\begin{figure*}[t]
    \setlength{\abovecaptionskip}{3pt}
    \setlength{\belowcaptionskip}{-0.0cm}
    \centering
    \includegraphics[width=\linewidth]{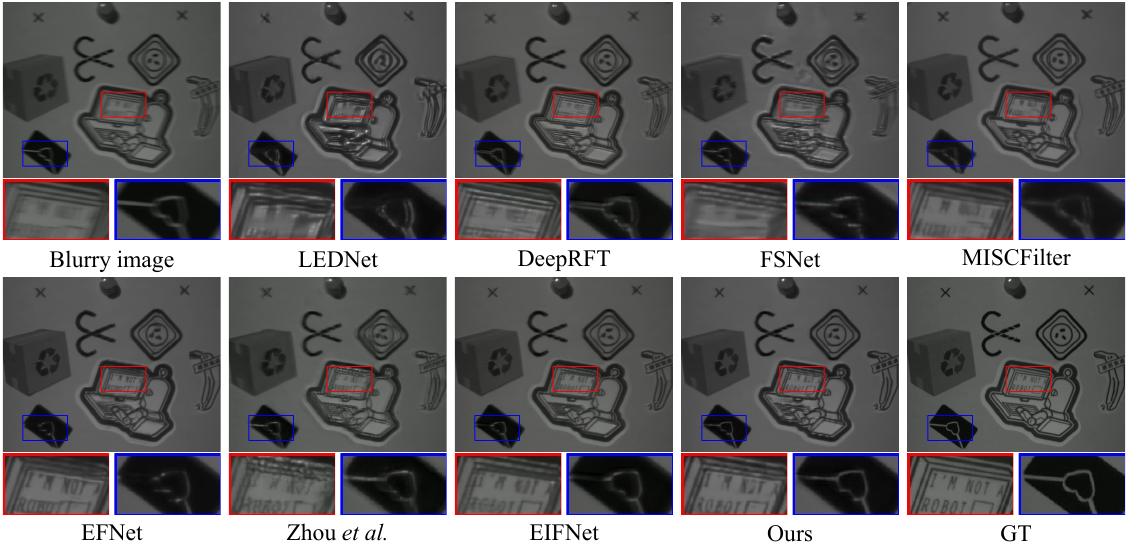}
    \caption{Comparison results of our deblurring method with others on the CSDD dataset.}
    \label{fig:deblur_real}
\end{figure*}

\begin{figure*}[t]
    \setlength{\abovecaptionskip}{3pt}
    \setlength{\belowcaptionskip}{-0.2cm}
    \centering
    \includegraphics[width=\linewidth]{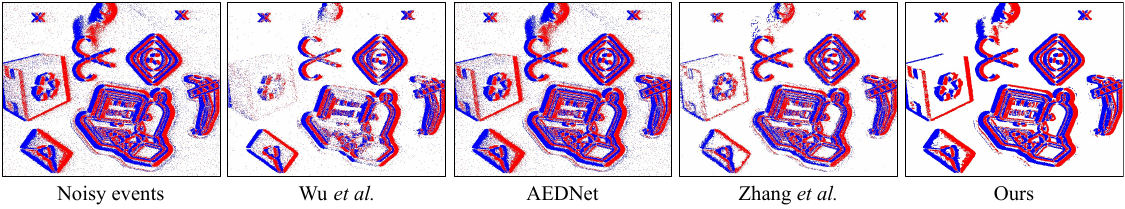}
    \caption{Comparison results of our denoising method with others on the CSDD dataset.}
    \label{fig:denoise_real}
\end{figure*}
\subsection{Implementation Details and Metrics}
We create a synthetic dataset based on the visible-nir dataset DVD \cite{jin2022darkvisionnet} for network training and evaluation. Specifically, we first generate visible and NIR image sequences according to previous work \cite{zhou2023deblurring}, and then construct blurry-sharp pairs of NIR images. Additionally, we use the open-source event simulator (ESIM) \cite{rebecq2018esim} to generate event streams corresponding to the visible images, and we create paired noise-clean event streams by introducing noise. We convert the event stream into a voxel grid representation and set the bin size of event voxels to $13$. Furthermore, we use captured CSDD datasets to verify the performance of the proposed method in real-world scenarios.

We set the batch size to 16 and train for $200$ epochs. We utilize the Adam optimizer \cite{kingma2014adam} with an initial learning rate of $0.0005$. The learning rate remains unchanged for the first $100$ epochs and gradually decreases to $0$ over the next $100$ epochs. Patches at the size of \(128\times128\) are randomly cropped from training samples, and random horizontal or vertical flips are applied. All experiments were conducted on a server using Python 3.8.0, PyTorch 1.12.0, and NVIDIA GeForce RTX 3090. The weighting factors \(\lambda_1\), \(\lambda_2\), and \(\lambda_3\) in Eq. \ref{eq:loss_total} are set to $1$, $1$, and $0.1$, respectively. We use structural similarity (SSIM) and peak signal-to-noise ratio (PSNR) as performance metrics for motion deblurring and use root mean square error (RMSE) as performance metrics for event denoising \cite{zhang2024neuromorphic}.

\subsection{Comparison of Motion Deblurring}
\begin{table}[t]
    \caption{Quantitative comparisons of the proposed MDEDNet to the state-of-the-art motion deblurring methods on the synthetic and CSDD datasets. The best performance is in bold.}
    \label{tab:deblur}
    \setlength{\tabcolsep}{1mm}
    \renewcommand\arraystretch{1.2}    
    \begin{center}
    \footnotesize
    \begin{tabular}{ccccccc}  
    \toprule
    \multirow{2}{*}{Method}  & \multirow{2}{*}{Event} & Params & \multicolumn{2}{c}{ \underline{\quad Synthetic \quad}} & \multicolumn{2}{c}{\underline{\quad\quad CSDD \quad\quad}} \\ 
    & & (M) &PSNR &SSIM &PSNR &SSIM  \\
    \midrule    
    LEDNet \cite{zhou2022lednet}      & \XSolidBrush &7.4&34.07 &0.8526 &27.58 &0.8686  \\
    DeepRFT \cite{mao2023intriguing}  & \XSolidBrush &9.6&35.67 &0.9074 &28.23 &0.8753 \\
    FSNet \cite{cui2023image}         & \XSolidBrush &14.8&35.35 &0.9040 &26.04 &0.8388  \\
    MISCFilter \cite{liu2024motion}   & \XSolidBrush &17.1&35.45 &0.9053 &28.38 &0.8821 \\
    \midrule      
    EFNet \cite{sun2022event}         & \Checkmark   &8.5 &36.90 &0.9134 &34.64 &0.9531\\
    Zhou \textit{et al.} \cite{zhou2023deblurring}   & \Checkmark   &11.8&34.88 &0.8955 &32.99 &0.9313 \\    
    EIFNet \cite{yang2023event}       & \Checkmark   &10.8&36.97 &0.9227&34.89 &0.9551  \\
    Ours                              & \Checkmark   &\textbf{4.8}&\textbf{37.79} &\textbf{0.9248} &\textbf{35.80} &\textbf{0.9618}  \\ 
    \bottomrule
    \end{tabular}
    \end{center}
\end{table}
We first evaluate the performance of our MDEDNet on the motion deblurring task. Both qualitative and quantitative comparisons are made with state-of-the-art frame-based methods, including LEDNet \cite{zhou2022lednet}, DeepRFT \cite{mao2023intriguing}, FSNet \cite{cui2023image}, and MISCFilter \cite{liu2024motion}, as well as event-based methods, including EFNet \cite{sun2022event}, Zhou \textit{et al.} \cite{zhou2023deblurring}, and EIFNet \cite{yang2023event}. Their official code is used and retrained to achieve optimal performance. Table \ref{tab:deblur} shows that our method outperforms other methods in deblurring both synthetic and CSDD datasets. Specifically, on the synthetic dataset, our method improves performance by \SI{2.12}{dB} over the best frame-based methods and by \SI{0.82}{dB} over the best event-based methods. Additionally, Figs. \ref{fig:deblur_syn} and \ref{fig:deblur_real} present qualitative visual comparisons on synthetic and CSDD datasets, respectively. Overall, these visual comparisons indicate that our method can recover sharper texture details and demonstrate strong generalization performance on real scenes.

\subsection{Comparison of Event Denoising}
\begin{table}[t]
    \caption{Quantitative comparisons of the proposed MDEDNet to the state-of-the-art event denoising methods on synthetic dataset. The best performance is in bold.}
    \label{tab:denoise}
    \setlength{\tabcolsep}{1mm}
    \renewcommand\arraystretch{1.2}    
    \begin{center}
    \footnotesize
    \begin{tabular}{ccccc}  
    \toprule
     & Wu \textit{et al.} \cite{wu2020probabilistic}  & AEDNet \cite{fang2022aednet} & Zhang \textit{et al.} \cite{zhang2024neuromorphic}& Ours \\
     \midrule
     RMSE&0.1896 & 0.2042& 0.1977&\textbf{0.1159}  \\
    \bottomrule
    \end{tabular}
    \end{center}
\end{table}
Furthermore, we compare MDEDNet with state-of-the-art event denoising methods, including Wu \textit{et al.}~\cite{wu2020probabilistic}, AEDNet \cite{fang2022aednet}, and Zhang \textit{et al.}~\cite{zhang2024neuromorphic}. Table \ref{tab:denoise} shows that the proposed method achieves the lowest RMSE on the synthetic dataset. Figs. \ref{fig:denoise_syn} and \ref{fig:denoise_real} illustrate the qualitative results on the synthetic and CSDD datasets. The results indicate that, compared to other state-of-the-art methods, the proposed method demonstrates a significant advantage in event denoising performance.

\begin{table}[t]
    \setlength{\abovecaptionskip}{-3pt}
    \setlength{\belowcaptionskip}{-0.0cm}
    \centering
    \caption{Ablation study of SCE and CMI modules on the synthetic dataset. \(\rm CMI_{j}\) represents the cross-modal interaction based on the \(j\)-th order gradient. The best performance is in bold.}
    \label{tab:ablation_module}
    \renewcommand\arraystretch{1.2}
    \begin{center}
    \footnotesize
    \begin{tabular}{cccccccc}  
    \toprule
    Case & SCE & \(\rm CMI_{0}\) &\(\rm CMI_{1}\) &\(\rm CMI_{2}\) & PSNR  & SSIM  & RMSE\\  
    \midrule
     \#1 &           &           &           &           &36.77           &0.9139   &0.1188\\
     \#2 &\Checkmark &           &           &           &37.20           &0.9188   &0.1185 \\
     \#3 &\Checkmark &\Checkmark &           &           &37.62           &0.9217   &0.1171\\  
     \#4 &\Checkmark &\Checkmark &\Checkmark &           &37.73           &0.9230   &0.1164 \\ 
     \#5 &\Checkmark &\Checkmark &\Checkmark &\Checkmark &\textbf{37.79}  &\textbf{0.9248} &\textbf{0.1159}  \\  
    \bottomrule
    \end{tabular}
    \vspace{-0.5ex} 
    \end{center}
\end{table}

\begin{figure}[h!]
    \setlength{\abovecaptionskip}{2pt}
    \setlength{\belowcaptionskip}{-0.3cm}
    \centering
    \includegraphics[width=0.98\linewidth]{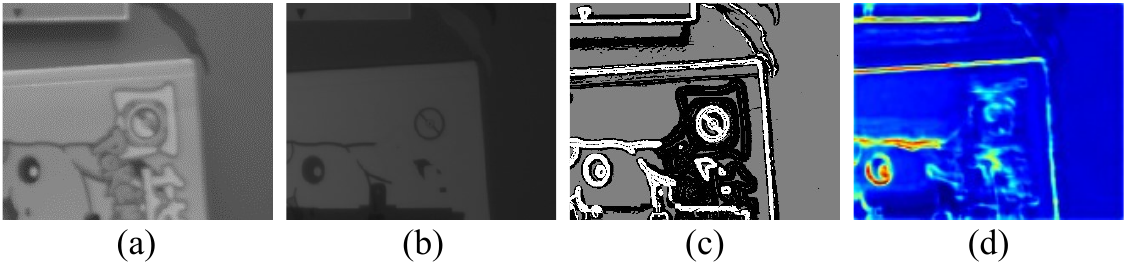}
    \caption{(a) and (b) are the visible light and NIR responses of the same scene. (c) The structural consistency \(C\) between different spectral responses. (d) The predicted spectral consistency feature map \(M^c\). }
    \label{fig:consistency_map_analysis}
\end{figure}

\subsection{Ablation Study}

\noindent\textbf{SCE Module.}
The SCE module allocates pixel-level weights by predicting the spectral consistency map \( M^c \). As shown in Table \ref{tab:ablation_module}, the introduction of the SCE module increases the PSNR of MDEDNet by \SI{0.43}{dB}. This result highlights the significant role of the SCE module in improving the deblurring performance. Furthermore, Fig. \ref{fig:consistency_map_analysis} shows the spectral response and the spectral consistency map \( M^c \) generated by the SCE module. The visualization results show that the SCE module effectively identifies inconsistent structures in the scene, thereby significantly enhancing the common features in both NIR and event data.

\noindent\textbf{CMI Module.}
The CMI module aims to achieve deep fusion between NIR and event features. As shown in Table \ref{tab:ablation_module}, the PSNR of MDEDNet decreases by \SI{0.59}{dB} without the CMI module. This result indicates that the CMI module plays a significant role in the collaborative enhancement of the two modalities. Furthermore, we analyze the contribution of different gradient information to cross-modal interaction. As shown in Table \ref{tab:ablation_module}, introducing first-order and second-order gradients effectively improves the deblurring and denoising performance of MDEDNet.

\noindent\textbf{Loss Functions.}
\begin{table}[t]
    \caption{Ablation study of loss functions on the synthetic dataset. The best performance is in bold.}
    \label{tab:ablation_loss}
    \renewcommand\arraystretch{1.2}
    \begin{center}
    \footnotesize
    \begin{tabular}{ccccccc}  
    \toprule
    Case &  \(L_{md}\) &  \(L_{ed}\) & \(L_{sc}\)   & PSNR  & SSIM  & RMSE       \\  
    \midrule
     \#1 &\Checkmark &           &           & 37.35          &0.9206         & ---    \\
     \#2 &           &\Checkmark &           & ---            &---             & 0.1403  \\ 
     \#3 &\Checkmark &\Checkmark &           & 37.59          & 0.9228         & 0.1261 \\  
     \#4 &\Checkmark &\Checkmark &\Checkmark &\textbf{37.79}  &\textbf{0.9248} & \textbf{0.1159}\\       
    \bottomrule
    \end{tabular}
    \end{center}
\end{table}
For the motion deblurring task, it is observed from Table \ref{tab:ablation_loss} that removing \(L_{ed}\) and \(L_{sc}\) leads to a degradation in reconstruction results. This is because \(L_{ed}\) introduces high-quality event streams, providing valuable spatiotemporal references for NIR deblurring. For the event denoising task, removing \(L_{md}\) and \(L_{sc}\) results in a decrease in denoising metrics, indicating that \(L_{md}\) and \(L_{sc}\) contribute to recovering noise-free events. \(L_{sc}\) effectively constrains spectral consistency, enabling highly relevant features for interactions between the NIR and event data. 

\section{Conclusion }
\label{sec:conclusion}
In conclusion, this paper introduced a novel framework to address motion blur in NIR imaging and severe noise in neuromorphic imaging under low-light conditions. To tackle these issues, we developed a synergistic neuromorphic imaging framework for jointly achieving NIR image deblurring and event denoising. Empirical evidence demonstrates that our spectral consistency enhancement significantly promotes cross-modal interaction and enhances imaging performance. This design effectively mitigates the gap caused by cross-modal differences and improves imaging capability in extremely dark scenes. Furthermore, to evaluate the synergistic neuromorphic imaging method, this study collected the NIR and event stream dataset, specifically tailored for low-light environments and varying system settings, further underscores our contribution by providing a robust platform for testing and benchmarking low-light NIR and event-based algorithms. Our scheme provides an enlightening reference for cross-modal tasks, highlighting its potential for challenging applications.


\section{supplementary Material}

In the supplementary material, we provide additional experimentation details, further analysis of spectrum and spectral consistency enhancement (SCE), further ablation experiments on cross-modal multi-order interaction (CMI), and additional qualitative results. 

We further provide a supplementary video to show our experimental platform, cross-spectral deblurring and denoising (CSDD) dataset, and results for both synthetic and real-world scenes.

\subsection{Additional Experimentation Details}
\subsubsection{Network Architecture}
To achieve joint optimization for NIR motion deblurring and event denoising, we design a dual-branch subnetwork. Each branch adopts a multi-scale encoder-decoder structure. The encoder comprises three encoding layers, with each layer containing two Res Blocks. The encoding process can be represented as
\begin{align}\label{eq:encode}
F_b = \mathcal{E}_b(B),  \quad F_e = \mathcal{E}_e( E_n ),
\end{align}
where \(B\) and \(E_n\) represent blurry NIR image and noisy events respectively. \(\mathcal{E}_b\) and \(\mathcal{E}_e\) denote the image encoder and event encoder, respectively. \(F_b\) and \(F_e\) represent the encoded features of image and event. Unlike single-task approaches, the proposed cooperative task requires fully leveraging the complementary advantages of both modalities. Therefore, we perform an efficient interactive fusion of the NIR-encoded and event-encoded features. The fusion process can be expressed as
\begin{align}\label{eq:fusion}
F_b^i, F_e^i = \rm CMI\big(\rm SCE(F_{b}, F_{e})\big),
\end{align}
where \(F_b^i\) and \(F_e^i\) represent the image interaction features and event interaction features, respectively. \(\rm SCE\) denotes the spectral consistency enhancement module, and \(\rm CMI\) signifies cross-modal multi-order interaction. The interaction features are then passed to the decoder, generating sharp images and clean events respectively. The structure of the decoder is similar to that of the encoder. The decoding process can be represented as
\begin{align}\label{eq:decode}
\widehat {S} = \mathcal{D}_b(F_b+F_e^i), \quad \widehat {E}_{c} = \mathcal{D}_e(F_e+F_b^i),
\end{align}
where \(\mathcal{D}_b\) and \(\mathcal{D}_e\) denote the image decoder and event decoder, respectively. \(\widehat {S}\) and \(\widehat {E}_{c}\) represent predicted sharp NIR image and clean events respectively.
\subsubsection{Synthetic Dataset Generation Pipeline}
The generation process of the synthetic dataset generally follows the method described in \cite{zhou2023deblurring}, with necessary adjustments and optimizations based on the unified tasks of NIR deblurring and event denoising. The Dark Vision Dataset \cite{jin2022darkvisionnet} is chosen as the ideal data source for constructing this synthetic dataset due to its inclusion of noise-free, paired visible light and NIR images. Specifically, for each scene, the synthetic dataset generation process can be described by the following steps:

\begin{itemize}
    \item Randomly crop the source image to \(320 \times 256\) pixels to obtain a sharp NIR image \(S\) and its corresponding visible light image;
    \item First, the Sobel filter is applied to both the visible light and NIR images for edge detection. Then, edge binarization is performed to calculate the structural consistency \(C\);
    \item The camera motion trajectory is randomly generated using the algorithm proposed in \cite{boracchi2012modeling};
    \item The visible light and NIR images are moved along pixel-level trajectories, resulting in an image sequence containing multiple potential frames during the exposure period;
    \item The event simulator (ESIM) \cite{rebecq2018esim} is used to generate the corresponding noise-free events \(E_c\) from the visible light image sequence. Simultaneously, the NIR image sequence undergoes averaging to produce the blurry NIR image \(B\);
    \item Noise is added to the noise-free event \(E_c\) according to the Poisson process to generate noisy events \(E_n\), and noise is also added to the blurry NIR image \(B\) to make it more representative of actual observations in real-world scenes \cite{lv2021attention}.
\end{itemize}
For each scene, we randomly generate \(20\) different camera motion trajectories. The final training dataset contains \(4000\) different images, and the test dataset contains \(600\) different images.
\subsubsection{CSDD Dataset Calibration}
\begin{figure}[t]
    \setlength{\abovecaptionskip}{3pt}
    \setlength{\belowcaptionskip}{-0.0cm}
    \centering
    \includegraphics[width=1\linewidth]{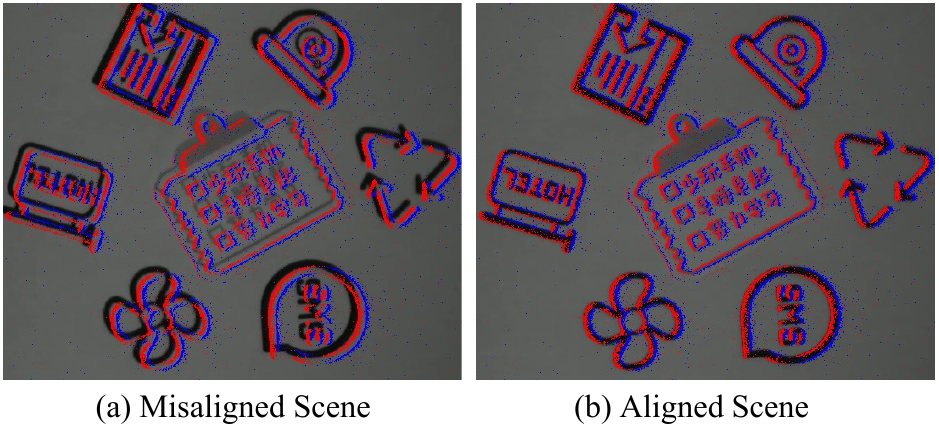}
    \caption{Alignment of NIR image with events. Aggregated events (red is positive, and blue is negative) are overlain with the NIR image.}
    \label{supplfig:aligned_scene}
\end{figure}
Similar to previous work \cite{zhou2023deblurring}, we adopt an approximate temporal synchronization strategy, writing a script to simultaneously trigger the capturing programs of the two cameras. We then estimate the homography matrix between the two cameras by manually selecting feature points, achieving geometric alignment of the NIR image sequences with the event stream. Fig. \ref{supplfig:aligned_scene} illustrates a typical example of the calibration results.

\subsection{Further Analysis on Spectrum and SCE}
\begin{figure*}[t]
    \setlength{\abovecaptionskip}{3pt}
    \setlength{\belowcaptionskip}{-0.0cm}
    \centering
    \includegraphics[width=1\linewidth]{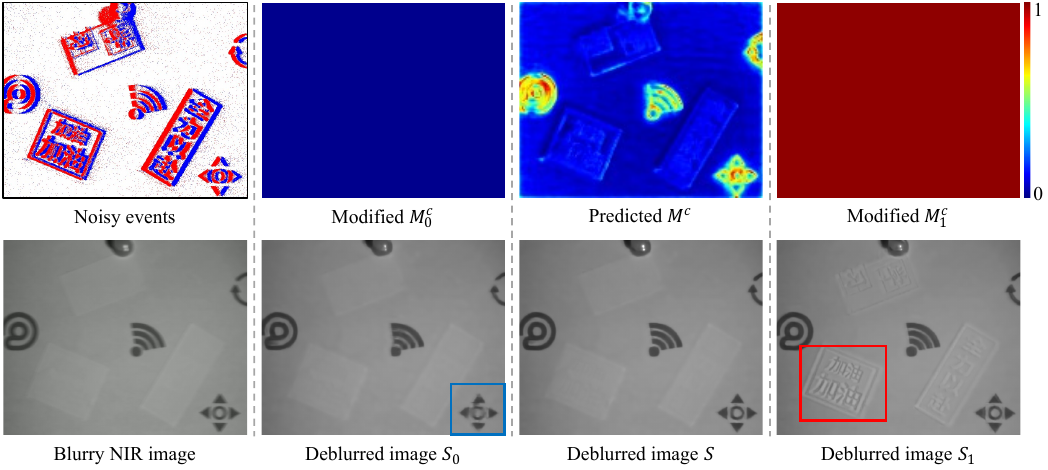}
    \caption{Deblurred results corresponding to \(M^c\) with different values. The SCE module can effectively identify consistent structures in the scene and generate the spectral consistency map \(M^c\). When \(M^c\) is set to \(0\), unexpected artifacts appear in the deblurred image \( S_0 \). When \(M^c\) is set to \(1\), the deblurred image \( S_1 \) effectively integrates information that was not originally present in the NIR image.}
    \label{supplfig:sce_analysis}
\end{figure*}
NIR and visible light primarily rely on the reflective properties of objects for information acquisition and analysis, which sharply contrasts with the working principle of long-wave infrared, based on the emissive properties of objects \cite{geng2024event}. Specifically, NIR and visible light exhibit significant similarities in many aspects, primarily due to their relative proximity in the spectral range and shared adherence to the physical mechanisms of reflection. However, despite these commonalities, the differences in wavelength range and the inherent variations in material reflectance can lead to inconsistent response characteristics of NIR and visible light under certain application scenarios. Therefore, when utilizing NIR cameras and event cameras for synergistic enhancement, it is necessary to thoroughly consider and distinguish their unique response characteristics.

The SCE module predicts and generates a spectral consistency map \(M^c\) by assessing the similarities and differences between NIR responses and event responses within a scene. The \(M^c\) assigns corresponding weight values to each pixel based on distinct structural features. As shown in the blue box area of Fig. \ref{supplfig:sce_analysis}, regions with consistent spectral responses receive higher weights, whereas those with inconsistent responses are assigned lower weights.

To further evaluate the influence of the SCE module on deblurring performance, we set all values in the \(M^c\) to zero, producing the corresponding deblurred image \( S_0 \). A \(M^c_0\) with all-zero values cannot provide pixel-level guidance for the fusion of NIR and event features, leading to undesired artifacts in \( S_0 \). In contrast, the \(M^c\) predicted by the SCE module accurately reflects the response differences between NIR and visible light in the scene. The deblurred image \( S \) demonstrates that the SCE module facilitates the recovery of sharp edges.

In addition, in certain application scenarios, it may be necessary to simultaneously integrate NIR and event information to achieve comprehensive reconstruction of both consistent and inconsistent regions. The SCE module effectively identifies differences in spectral responses within the scene, enabling subsequent fusion processes to adapt based on the \(M^c\). This adaptability implies that manually modifying the values of the \(M^c\) enables diverse reconstruction results. For instance, setting all values in the \(M^c\) to 1 signifies that both NIR and event responses are fully incorporated into the fusion. As illustrated in the red box area of Fig. \ref{supplfig:sce_analysis}, the deblurred image \( S_1 \) effectively integrates information that is originally absent in the NIR, thereby improving both the reconstruction quality and the richness of details.

\subsection{Further Ablation Experiments on CMI}
\begin{table}[t]
    \caption{Quantitative comparisons of different fusion methods on the synthetic dataset. \(\rm CMI_{j}\) represents the cross-modal interaction based on the \(j\)-th order gradient. The best performance is in bold.}
    \vspace{-2.5ex}
    \label{suppltab:cmi}
    \setlength{\tabcolsep}{1mm}
    \renewcommand\arraystretch{1.2}    
    \begin{center}
    \footnotesize
    \begin{tabular}{ccccccc}  
    \toprule
     Fusion & Add  & Concat. & Attention & \(\rm CMI_{0}\) & \(\rm CMI_{0,1}\) & \(\rm CMI_{0,1,2}\) \\
     \midrule
     PSNR&36.77  & 36.44  &37.12   &37.17   &37.28  &\textbf{37.39}  \\
     SSIM&0.9139 & 0.9136 &0.9143   &0.9145  &0.9177 &\textbf{0.9193} \\
    \bottomrule
    \end{tabular}
    \end{center}
\end{table}
To further validate the effectiveness of the CMI module, we compare it with several common fusion methods, including addition, concatenation, and cross-attention. As shown in Table \ref{suppltab:cmi}, fusing complementary information from two modalities through simple addition or concatenation has limited effectiveness, making it difficult to fully leverage the synergy between modalities. Although the cross-attention captures the relevant parts of features from both modalities, relying solely on shallow interactions between the image and event domains is insufficient to achieve optimal results. In contrast, CMI significantly enhances deblurring performance by fine-grained fusion of multi-order gradient features from images with event features, demonstrating its superiority in capturing cross-modal information.

\subsection{Additional Qualitative Results}
\subsubsection{Low-light Environment}
\begin{figure*}[t]
    \setlength{\abovecaptionskip}{3pt}
    \setlength{\belowcaptionskip}{-0.2cm}
    \centering
    \includegraphics[width=1\linewidth]{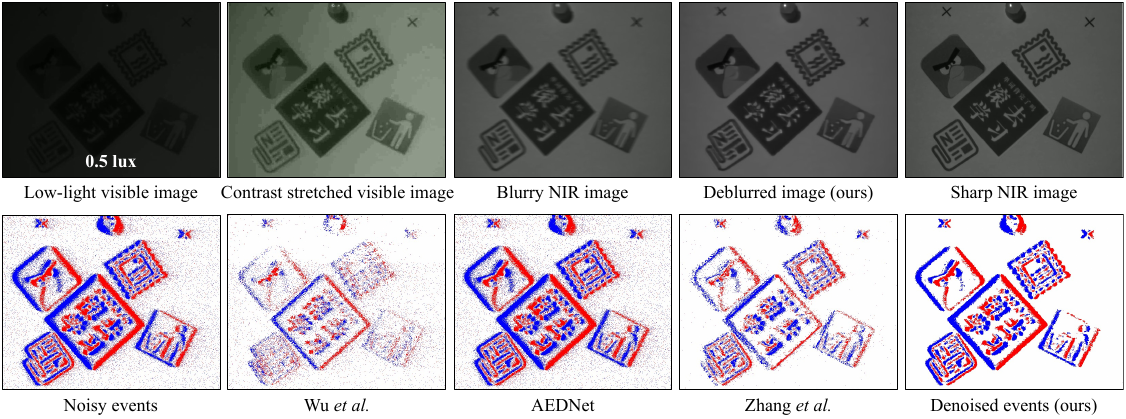}
    \caption{Comparison of visible light image, events, and NIR image in low-light environment, as well as the NIR deblurring and event denoising results of the proposed method. Our method effectively reconstructs sharp NIR image and clean events in challenging environment.}
    \label{supplfig:low_light}
\end{figure*}
Fig. \ref{supplfig:low_light} illustrates qualitative results of NIR deblurring and event denoising in low-light environment. Under an illumination condition of \(0.5\) lux, visible light image fail to convey clear scene information. Although contrast stretching enhances the visual clarity of the image, it simultaneously introduces significant noise. Event cameras, though highly sensitive in low-light environments, are prone to capturing substantial noise in their event data. In contrast, NIR image achieves a higher signal-to-noise ratio under such conditions but is susceptible to motion blur in dynamic scenes. In the unified task of NIR motion deblurring and event denoising, NIR cameras and event cameras demonstrate significant complementary advantages. Experimental results demonstrate that the proposed method effectively reconstructs sharp NIR image and high-quality, clean events in challenging environment.
\subsubsection{Synthetic Dataset}
\begin{figure*}[t]
    \setlength{\abovecaptionskip}{3pt}
    \setlength{\belowcaptionskip}{-0.0cm}
    \centering
    \includegraphics[width=1\linewidth]{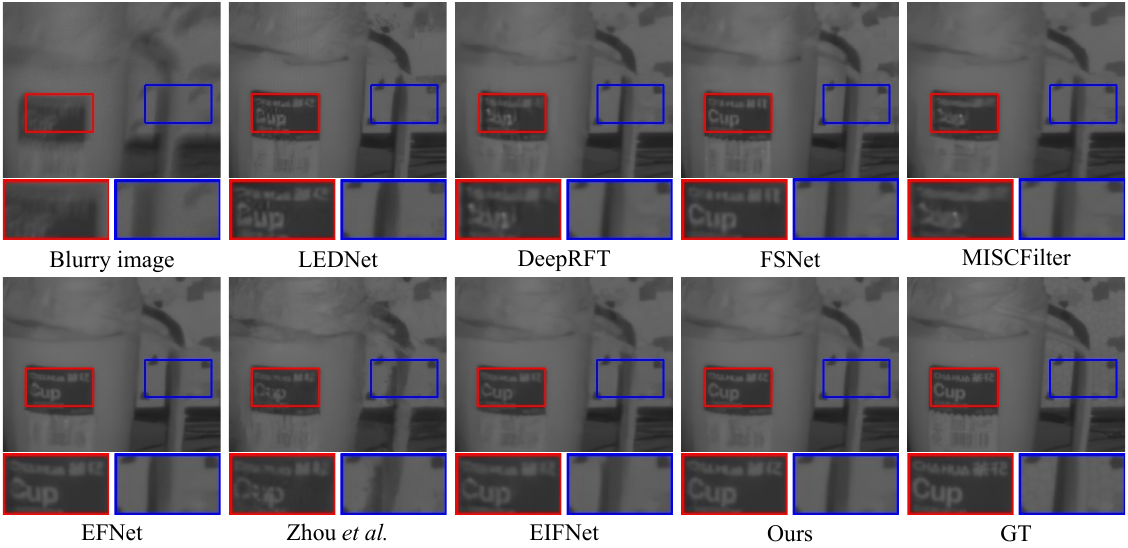}
    \caption{Comparison results of our deblurring method with others on the synthetic dataset.}
    \label{supplfig:deblur_syn_1}
\end{figure*}

\begin{figure*}[t]
    \setlength{\abovecaptionskip}{3pt}
    \setlength{\belowcaptionskip}{-0.2cm}
    \centering
    \includegraphics[width=1\linewidth]{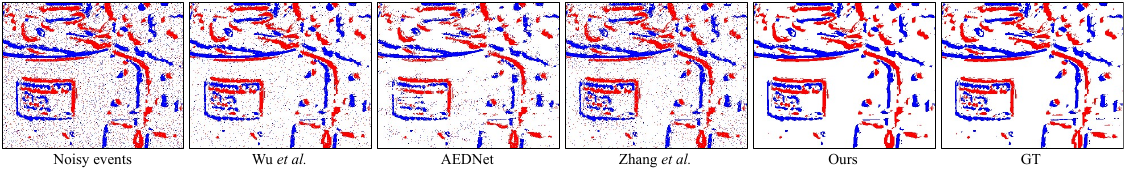}
    \caption{Comparison results of our denoising method with others on the synthetic dataset. The events come from the same scene as the NIR in Fig. \ref{supplfig:deblur_syn_1}.}
    \label{supplfig:denoise_syn_1}
\end{figure*}
Figs. \ref{supplfig:deblur_syn_1} and \ref{supplfig:denoise_syn_1} provide additional qualitative results for NIR deblurring and event denoising on the synthetic dataset, respectively. In the image deblurring task, frame-based deblurring methods face significant challenges when reconstructing sharp images directly from blurry ones in complex dynamic scenes. Although event-based deblurring methods show improvements in reconstruction quality, their results are often accompanied by artifacts and distortions due to event noise and spectral inconsistencies. In contrast, the proposed method effectively improves deblurring quality by introducing a spectral consistency enhancement module and a cross-modal multi-order interaction module. For the event denoising task, aided by high signal-to-noise ratio NIR images, the proposed method effectively suppresses event noise.

\subsubsection{CSDD Dataset}
\begin{figure*}[t]
    \setlength{\abovecaptionskip}{3pt}
    \setlength{\belowcaptionskip}{-0.0cm}
    \centering
    \includegraphics[width=0.86\linewidth]{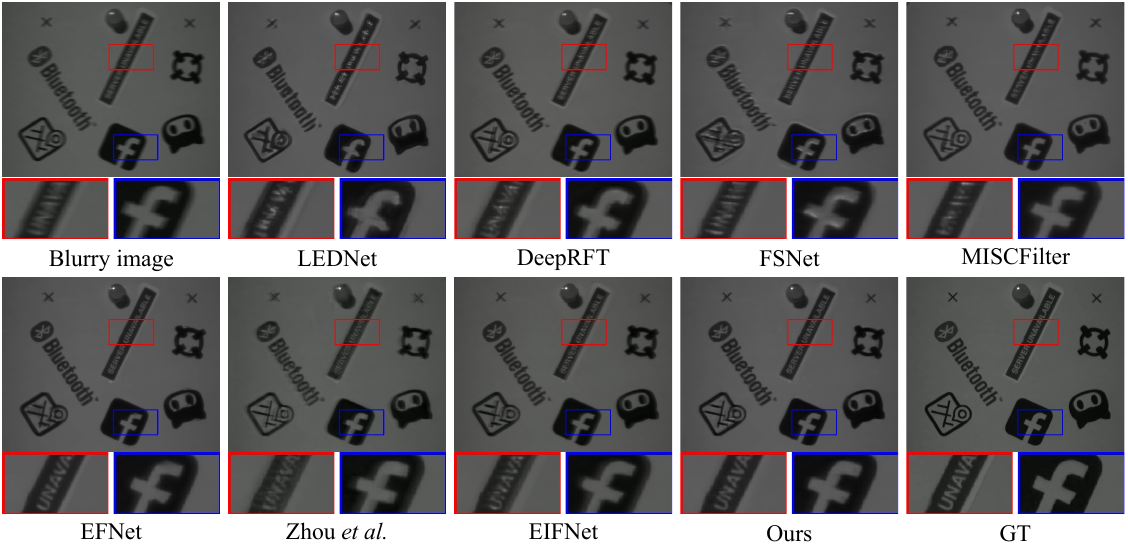}
    \includegraphics[width=0.86\linewidth]{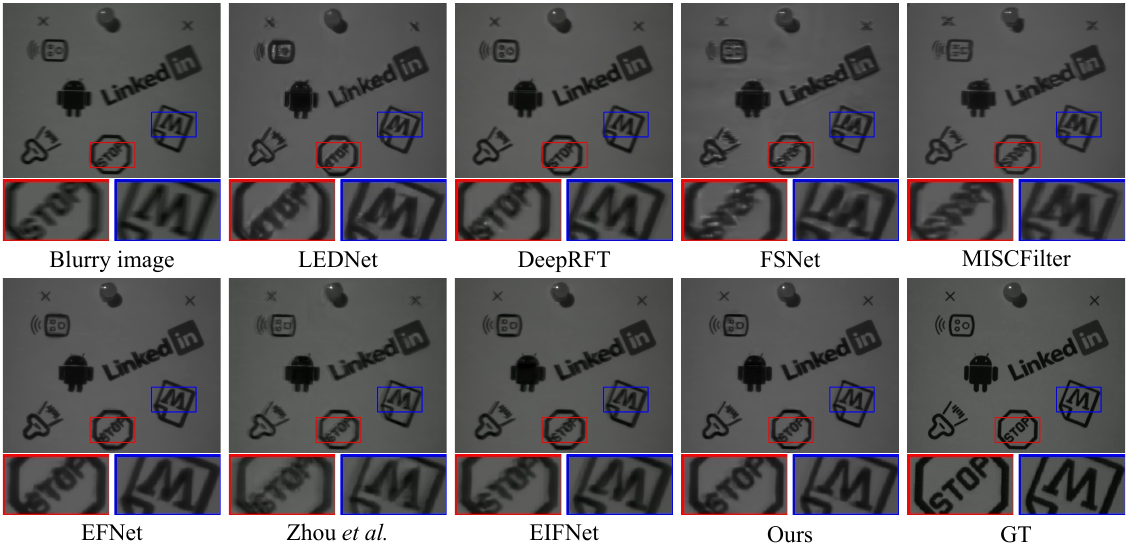}
    \caption{Comparison results of our deblurring method with others on the CSDD dataset. The scene does not contain spectral inconsistency for both NIR and events.}
    \label{supplfig:deblur_real_3_4}
\end{figure*}

\begin{figure*}[t]
    \setlength{\abovecaptionskip}{3pt}
    \setlength{\belowcaptionskip}{-0.2cm}
    \centering
    \includegraphics[width=0.86\linewidth]{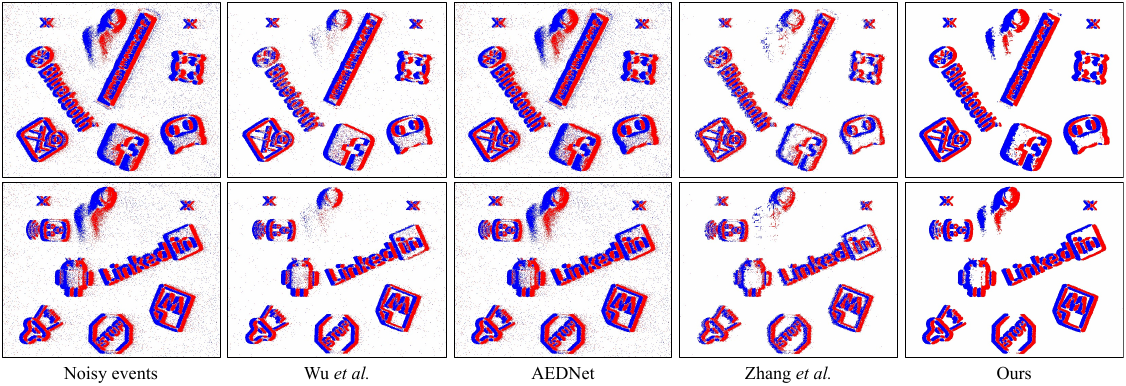}
    \caption{Comparison results of our denoising method with others on the CSDD dataset. The events come from the same scene as the NIR in Fig. \ref{supplfig:deblur_real_3_4}.}
    \label{supplfig:denoise_real_3_4}
\end{figure*}

\begin{figure*}[t]
    \setlength{\abovecaptionskip}{3pt}
    \setlength{\belowcaptionskip}{-0.0cm}
    \centering
    \includegraphics[width=0.86\linewidth]{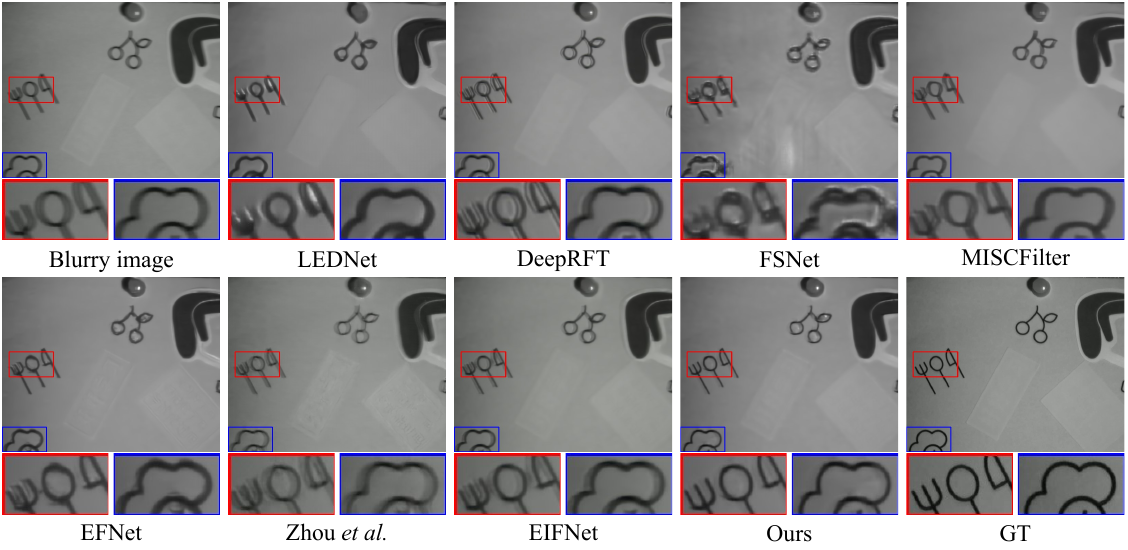}
    \includegraphics[width=0.86\linewidth]{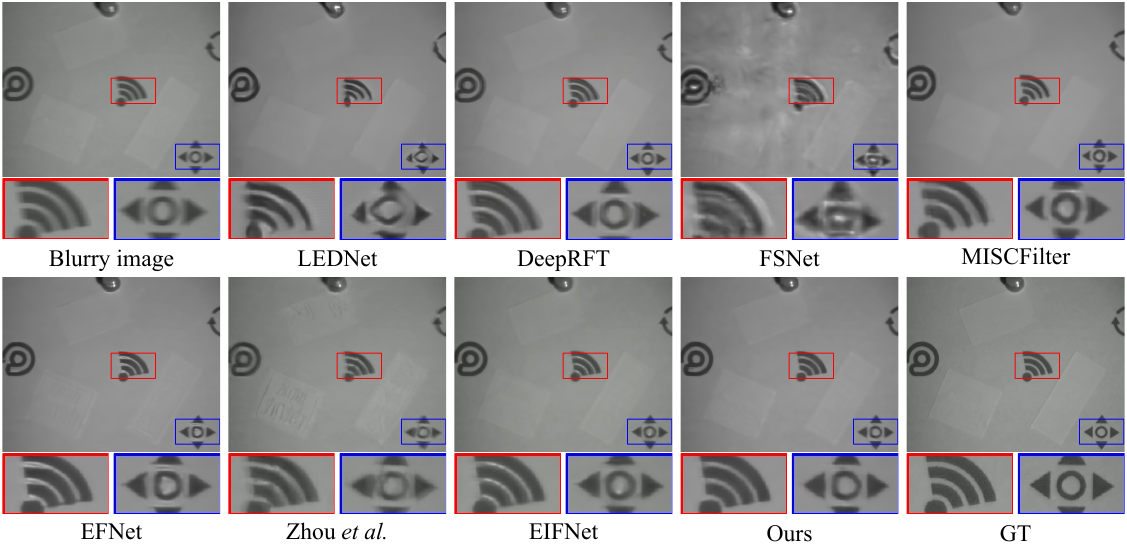}
    \caption{Comparison results of our deblurring method with others on the CSDD dataset. The scene contains spectral inconsistency for both NIR and events.}
    \label{supplfig:deblur_real_1_2}
\end{figure*}

\begin{figure*}[t]
    \setlength{\abovecaptionskip}{3pt}
    \setlength{\belowcaptionskip}{-0.2cm}
    \centering
    \includegraphics[width=0.86\linewidth]{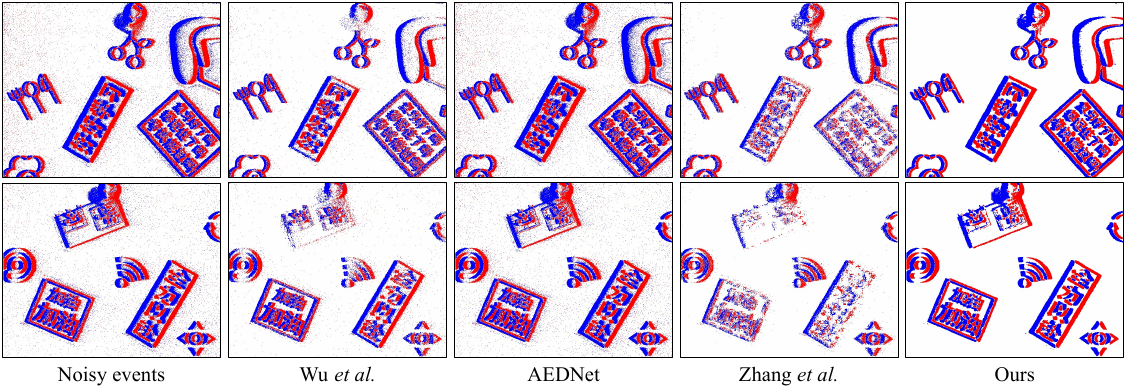}
    \caption{Comparison results of our denoising method with others on the CSDD dataset. The events come from the same scene as the NIR in Fig. \ref{supplfig:deblur_real_1_2}.}
    \label{supplfig:denoise_real_1_2}
\end{figure*}

Figs. \ref{supplfig:deblur_real_3_4} and \ref{supplfig:deblur_real_1_2} present additional qualitative results of NIR deblurring on the CSDD dataset, while Figs. \ref{supplfig:denoise_real_3_4} and \ref{supplfig:denoise_real_1_2} provide additional qualitative results of event denoising on the same dataset. Figs. \ref{supplfig:deblur_real_3_4} and \ref{supplfig:denoise_real_3_4}, as well as Figs.\ref{supplfig:deblur_real_1_2} and \ref{supplfig:denoise_real_1_2}, correspond to the same scenes with NIR responses and event responses, respectively. Specifically, Figs. \ref{supplfig:deblur_real_3_4} and \ref{supplfig:denoise_real_3_4} do not contain spectrally inconsistent structures, whereas Figs. \ref{supplfig:deblur_real_1_2} and \ref{supplfig:denoise_real_1_2} include spectrally inconsistent structures. These results demonstrate that, regardless of whether spectral inconsistency is present in the scene, the proposed method reconstructs sharp images from blurred NIR inputs and restores clean events from noisy ones. Compared with contrasting methods, the proposed method shows significant advantages in deblurring and denoising performance in real scenes.
\clearpage

\clearpage

\bibliographystyle{ieeetr}
\bibliography{arxiv}

\end{document}